# Analyzing the Impact of COVID-19 on Economy from the Perspective of User's Reviews

Fatemeh Salmani
Department of Computer Engineering
University of Birjand
Birjand, Iran
Salmani_fatemeh98@birjand.ac.ir

Hamed Vahdat-Nejad[*]
Department of Computer Engineering
University of Birjand
Birjand, Iran
vahdatnejad@birjand.ac.ir

Hamideh Hajiabadi
Department of Computer Engineering
Birjand University of Technology
Birjand, Iran
hajiabadi@birjandut.ac.ir

*Abstract*— One of the most important incidents in the world in 2020 is the outbreak of the Coronavirus. Users on social networks publish a large number of comments about this event. These comments contain important hidden information of public' opinion regarding this pandemic. In this research, a large number of Coronavirus-related tweets are considered and analyzed using natural language processing and information retrieval science. Initially, the location of the tweets is determined using a dictionary prepared through the Geo-Names geographic database, which contains detailed and complete information of places such as city names, streets, and postal codes. Then, using a large dictionary prepared from the terms of economics, related tweets are extracted and sentiments corresponded to tweets are analyzed with the help of the RoBERTa language-based model, which has high accuracy and good performance. Finally, the frequency chart of tweets related to the economy and their sentiment scores (positive and negative tweets) is plotted over time for the entire world and the top 10 economies. From the analysis of the charts, we learn that the reason for publishing economic tweets is not only the increase in the number of people infected with the Coronavirus but also imposed restrictions and lockdowns in countries. The consequences of these restrictions include the loss of millions of jobs and the economic downturn.

I. INTRODUCTION

Nowadays, a lot of comments are published by users on social networks, especially Twitter. By analyzing these comments, we can understand the public's opinion about the events and achieve valuable information. For this purpose, reviews related to the subject should be extracted from the social media and various processes (Such as sentiment analysis to identify the preferences of tourists [1, 2] and extracting food preferences of users [3]) should be performed on them.

The outbreak of Coronavirus has had a great impact on various areas such as health, education, economy, and agriculture [4]. Due to the effects that the outbreak has on the economy, several analyses have been performed in this field. These analyses include the topic modeling and sentiment analysis to investigate the five dominant issues during the Covid-19 outbreak [5], and financial sentiment analysis influence of Covid-19 on the global stock market indices [6]. Rahman et al. [7] have investigated tweets related to 51 states to analyze the sentiment of people about the reopening of the US economy. Then, socioeconomic characteristics (such as household income and education), and environmental characteristics (such as population density), and the number of cases of Covid-19 were analyzed. Finally, the positive and negative sentiments of individuals about the reopening of the economy are identified. The study has yielded interesting results, including the agreement to reopening, from the viewpoint of those with lower levels of education and income, and the opposition of those with higher levels of income and education [7].

In this paper, we analyze the sentiment of economic tweets related to Covid-19 and analyze the views of users in different countries. Initially, millions of tweets have been used from March 23 to June 23, 2020. Tweets are content-analyzed both spatially and economically. To determine the location of tweets, a list of names of countries, and provinces/states have been prepared in the form of a dictionary using the Geo-Names[1] geographical database. This database provides us with complete and accurate information about the names of places that can be used to tag the location of tweets. Then, using the Wordnet[2] ontology, and the Oxford[3] dictionary, the words related to the field of economics along with the names of important banks in each country and the currency of countries, have been prepared in the form of a comprehensive dictionary. Then, using this dictionary, tweets related to economics are extracted. Afterward, the sentiments of tweets are analyzed using the recent RoBERTa [8] language-based model, which has better accuracy and performance than similar models. Finally, trends of the frequency of tweets and sentiment analysis (positive and negative tweets) are drawn for the whole world, and the top ten[4] economies. From the analysis of the charts, we learn that the reason for posting economic tweets is not only due to the increase in the number of people infected with the Coronavirus but also imposed restrictions and lockdowns in countries. The consequences of these restrictions include the loss of millions of jobs and the economic downturn.

---

[1] https://www.geonames.org/
[2] http://wordnetweb.princeton.edu/perl/webwn
[3] https://www.oxfordreference.com/view/10.1093/acref/9780199237043.001.0001/acref-9780199237043
[4] https://howmuch.net/articles/the-world-economy-2019

The rest of the paper is structured as follows: In the second section, the proposed method is stated. In the third section, the evaluation is presented. Finally, in the fourth section, the conclusions and next directions of research are discussed.

## II. PROPOSED METHOD

The purpose of this study is to analyze Coronavirus-related tweets to extract the sentiments of users about the economy over time. First, tweets related to Covid-19 and economics are extracted. Then sentiments of the tweets are analyzed.

### A. Tweet extraction and tagging

The outbreak of Covid-19 began in late 2019 when initially only China hosted the virus, but it began to spread all over the word in February of 2020. We consider the English tweets published from March to June of 2020. The keywords "Corona", "Coronavirus", "Covid", "pandemic", "sarscov2" and "Covid-19" have been used to extract Coronavirus-related tweets.

Next, we aim to specify the location of each tweet. To tag locations, we have compiled a list of place names that include country, states/provinces, and city names using the Geo-Names geographic database. This database provides accurate and complete information of places such as city names, streets, and even postcodes. This list, which has about 7,000 words, is then used as a Gazetteer list in a GATE [9] pipeline. Finally, each tweet's words are matched to the list. For each city, state/province, or country name mentioned in the tweet, the name of the country is attributed to the tweet as a feature.

We have collected economics-related terms as a dictionary, through Oxford Dictionary, Wordnet ontology, and with the help of an expert. Figure 1 shows the word cloud from the prepared economic dictionary. We have also included the names of banks, stock companies, and various currencies in our dictionary. Then, like the location tagging method, the relevant tweets are investigated via a GATE software pipeline. In this way, the dictionary, which has about 500 words, is used as a Gazetteer list in a GATE pipeline and the tweets' words are matched with it. If there is an economic term in the tweet, the economy label is assigned to that tweet.

### B. Sentiment analysis of economic tweets

Sentiment analysis is one of the most important processes performed on tweets. By sentiment analysis, public opinion can be evaluated about important issues and events. In this paper, sentiment analysis is performed to examine and evaluate the thoughts and attitudes of people during the outbreak of Covid-19 about economics and related issues. We use the RoBERTa model [8] to analyze sentiments because of its high accuracy, speed, and performance. RoBERTa is the result of an improvement on the BERT[5] model. After calculating the sentiment of each tweet, the frequency of positive and negative tweets from the economic point of view is calculated. The frequency trend of economics and sentiment analysis is then plotted over time for the entire world and the top 10 economies. This analysis of economic tweets can greatly help governments learn about people's thoughts and feelings regarding economic issues and make the right decisions to improve the economic conditions of individuals and communities during pandemics.

Figure 1. Word cloud of the Dictionary of Economics

## III. EVALUATION

A dataset of more than 2 million Covid-19 related tweets on each Tuesday from March 23 to June 23, 2020, is used. To retrieve the location of each tweet, a list of place names (countries, states/provinces, and cities) has been prepared in the form of a comprehensive dictionary using the Geo-Names geographical database. This dictionary is stored as a location and used in a GATE [9] pipeline as a Gazetteer list to match the tweets with the list elements. With each occurrence of the name of a city or state/province or country, in the text of the tweet, the name of that country is assigned as the location feature. Finally, if the proposed sentiment analysis model gives the tweet a score of one, it is positive, and if it gives a score of zero, the tweet is negative.

Figure 2 shows the frequency of economic tweets, sentiment analysis results as well as statistics of the daily confirmed Covid-19 cases for the whole world. The vertical axis on the right represents the official statistics of Coronavirus cases and the vertical axis on the left represents the number of tweets, and the horizontal axis represents the 14 weeks. The official confirmed cases are obtained using the Coronavirus resource center of JOHNs Hopkins[6] university and medicine. In the seventh week, the curve of the frequency of tweets has a peak in which, the number of negative tweets is more than the number of positive tweets. A review of this week's tweets reveals that the main reason for the increase in tweets, in addition to the increase in the number of people suffering from Corona, is the loss of jobs and the severe economic record in various countries. In that week, the restrictions and lockdowns have become much more serious than before and have led to the closure of many guilds, companies, restaurants, etc. Figure 3 shows the same diagrams for 10 countries.

---

[5] https://huggingface.co/transformers/model_doc/roberta.html

[6] https://Coronavirus.jhu.edu/map.html

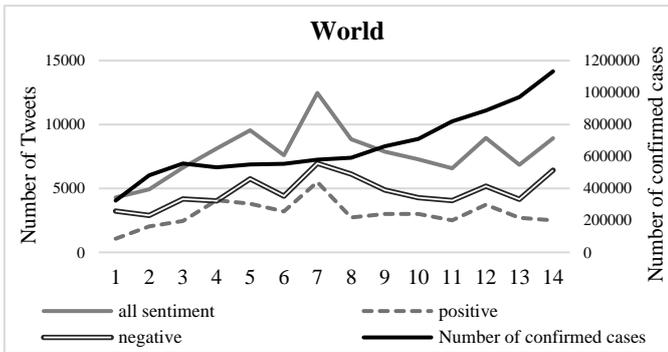
Figure 2. Trends of economic tweets, sentiment analysis, and official Covid-19 cases for the world

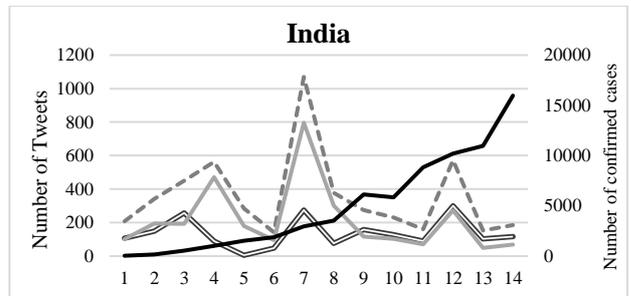

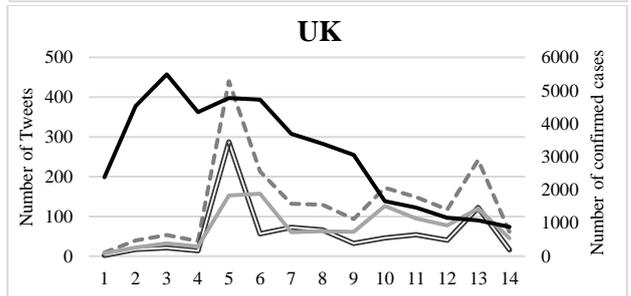

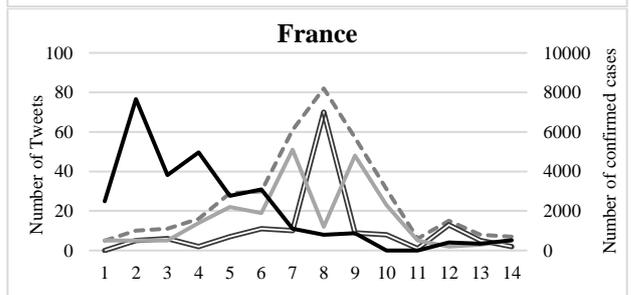

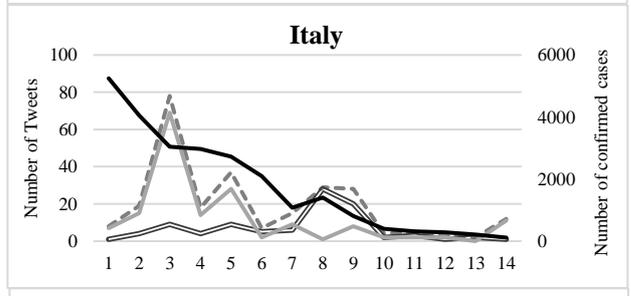

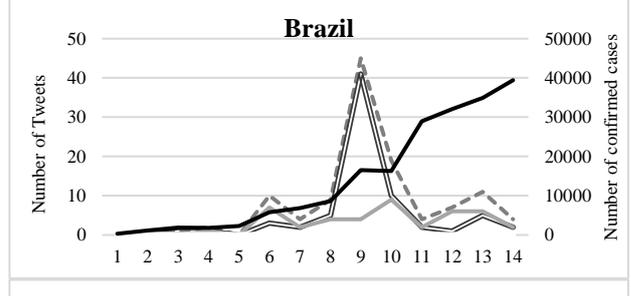

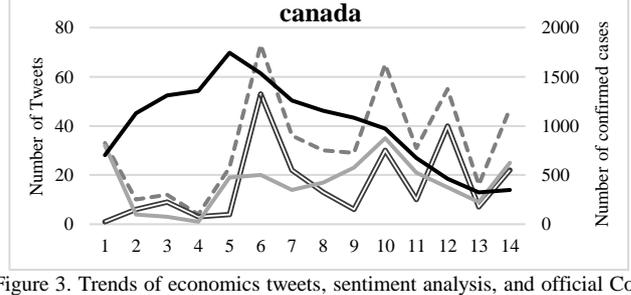

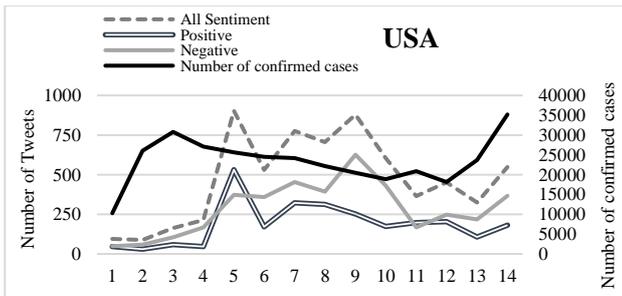

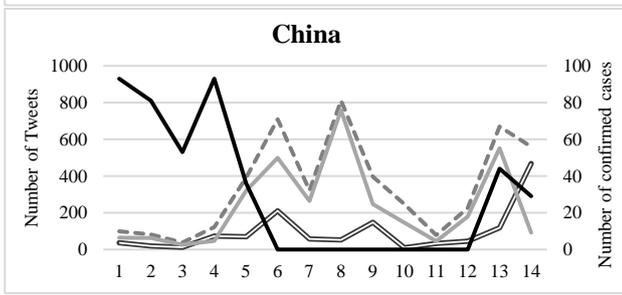

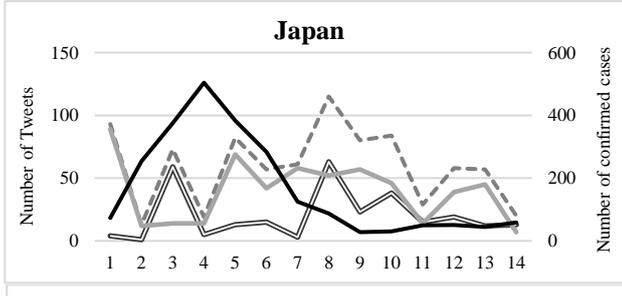

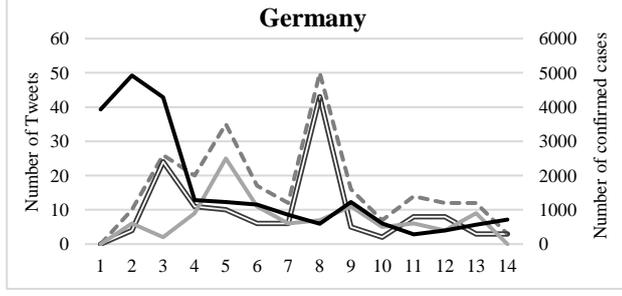

Figure 3. Trends of economics tweets, sentiment analysis, and official Covid-19 cases

Analyzing the results reveal that there is a correlation in several countries between the frequency of economics-related tweets and the official statistics of Coronavirus. Examples include Italy, the United Kingdom, Brazil, and Canada. Besides, in countries such as China, the United States, Germany, France, and Japan, in some weeks the official statistics of Coronavirus cases decrease, but the tweets related to the economy increase. The reason for this increase in tweets is due to government-related actions (including lockdowns and restrictions) and relations between countries during the Covid-19 outbreak.

For a closer look, we investigate the largest peaks in the frequency curves of tweets. For example, in the US diagram, there are two major peaks, corresponding to weeks' numbers 5 and 9. In the 5th week, the number of positive tweets is more than the number of negative tweets. One of the reasons for this positive peak is the government's ban on immigration to the United States to keep jobs. Also in the 9th Week, the American people protested against the government blaming the government for not taking the pandemic seriously and for losing 36 million jobs.

In China, although the official incidence rate drops to zero in weeks 6 to 12, two major peaks in weeks 6 and 8 are seen in the tweet frequency curves. The reasons for these peaks are the escalation of the economic war between China and the United States and the ban on Chinese exports and imports to Australia. Finally, an overview of these diagrams shows that in most countries, there are the largest peaks in the tweet frequency curves between weeks 6 and 10. The main reasons for these peaks are the restrictions imposed in each country, which have caused the recession, the sharp drop in the oil price, the loss of thousands of jobs, and the spread of unemployment.

## IV. CONCLUSION

This paper presents a method for analyzing Corona-related tweets to investigate the effect of the virus on the global economy from the perspective of users. Initially, Corona-related tweets have been randomly sampled every Tuesday in 14 weeks. The tweets were then tagged using a dictionary from the Geo-Name database and the GATE pipeline. Also, using a dictionary prepared from the terms of economics, the relevant tweets have been extracted and with the help of the RoBERTa model, the type of sentiment (positive and negative) of each tweet has been calculated. Finally, the frequency diagrams of economic tweets as well as sentiment analysis results are drawn for each country and the whole world. The analysis of the graphs has indicated that people in different countries are dissatisfied with the imposition of restrictions and lockdowns because they have resulted in a severe economic recession and the loss of millions of jobs.

This paper just focused on English language tweets. Also, analyzing Twitter data cannot be a testament to the thoughts and sentiments of the entire population because many people do not use Twitter. Given that a large number of tweets are written in languages other than English, providing a language-independent approach can provide a more complete and accurate analysis of public opinion views. Due to the different nature of languages, investigating non-English tweets such as Chinese, Arabic, Persian, Italian, etc. is future research work. Besides, tweets can be clustered according to their topics through topic modeling, and then each cluster can be analyzed.

ACKNOWLEDGMENT

The authors would like to thank "Mr. Parsa Bagherzadeh" for assisting in the implementation of this article.